\pdfoutput=1
\documentclass[a4paper]{article}
\usepackage{float}
\usepackage{graphicx}
\usepackage{caption}
\usepackage{subcaption}
\usepackage{gensymb}
\usepackage{bbm}
\usepackage{amsmath}
\usepackage{nth}
\usepackage{enumitem,kantlipsum}
\usepackage{hyperref}
\usepackage{url}
\usepackage[numbers]{natbib}
\usepackage{tabularx}
\usepackage[usestackEOL]{stackengine}
\usepackage[toc,page]{appendix}
\usepackage{xfrac}

\usepackage{geometry}
\newcommand{\OR}{ORCEA}
\newcommand{\M}{\mathsf{M}}
\newcommand{\MS}{\Phi_{\mathsf{M}}}
\newcommand{\m} {m}
\newcommand{\mm} {$m$ }
\newcommand{\MSp}{\underset{\MS}{p}}

\newcommand{\GM}{\mathbf{G}_m}
\newcommand{\GMev}{\GM(\ev)}
\newcommand{\GMevm}{$\GM(\ev)$ }
\newcommand{\GAM}{\mathbf{G}^0_m}

\newcommand{\ev}{\mathbbm{e}}
\newcommand{\evm}{$\mathbbm{e}$ }
\newcommand{\evS}{{\Phi_{\mathsf{e}}}}
\newcommand{\evSe}{\evS^\mathsf{e}}
\newcommand{\evSa}{\evS^\mathsf{a}}
\newcommand{\evp}{\underset{\evS}{p}}

\newcommand{\ES}{\mathbbm{E}}
\newcommand {\ESS}{\Phi_{\mathsf{E}}}
\newcommand{\ESC}{\mathbbm{C}}
\newcommand {\ESCS}{\Phi_{\ESC}}
\newcommand{\ESU}{\mathbbm{U}}
\newcommand {\ESUS}{\Phi_{\ESU}}

\newcommand{\ESp}{\underset{\ESS}{p}}
\newcommand{\ESCp}{\underset{\ESCS}{p}}
\newcommand{\ESUp}{\underset{\ESUS}{p}}

\newcommand {\obS}{\Phi_{\mathsf{\omega}}}

\newcommand{\ob}{{\omega}}

\newcommand{\GOB}{\mathbf{G_{\Omega}}}
\newcommand{\GOBo}{\mathbf{G_{\Omega(\ob)}}}

\newcommand{\HipS}{\mathbbm{H}}
\newcommand{\Hip}{\mathcal{H}}

\newcommand*\diff{\mathop{}\!\mathrm{d}}

\title{{\OR} \linebreak \large{ Object Recognition by Continuous Evidence Assimilation }}
\author{Oded Cohen}

\begin{document}

\maketitle

\begin{abstract}
This paper presents {\OR}, a novel object recognition method applicable for objects describable by a generative model. The primary goal of {\OR} is to maintain a probability density distribution of possible matches over the object parameter space, while continuously updating it with incoming evidence; detection and regression are by-products of this process.
{\OR} can project primitive evidence of various types (edge element, area patches etc.) directly on the object parameter space; this made possible by the study phase where {\OR} builds a probabilistic model, for each evidence type, that links evidence and the object-parameters under which they were created. The detection phase consists of building the joint distribution of possible matches resulting from the set of given evidence, including possible grouping to signal/noise; no additional algorithmic steps are needed, as the resulting PDF encapsulates all knowledge about possible solutions.
{\OR} represents the match distribution over the parameter space as a set of Gaussian distributions, each representing a concrete probabilistic hypothesis about the object, which can be used outside its scope as well.
{\OR} was tested on synthetic images with varying levels of complexity and noise, and shows satisfactory results; real-world input will be tested the next stage of the project.

\end{abstract}

\section{Introduction}
In this article I introduce {\OR}, a novel method for object detection, applicable for objects describable by generative model. This method bridges directly between low-level input (edge elements, color patches etc.) and the distribution of possible matches. This is done by projecting the evidence PDF into the object parameter space, hence rendering unnecessary nearly any algorithmic steps. 

{\OR} was developed based on my accumulative experience in the field of industrial object recognition, which has expanded over the years to include a large variety of object types, from musical notes to plant parts.
So far, {\OR} has only been tested on synthetic images; this article represents its theory, describes briefly its current implementation, and presents first test results on synthetic input. Future articles will deal with implementation issues, real world images and benchmarks.

\subsection{Object recognition from engineering perspective.}
I will start by posing three engineering questions relevant to physical system analysis, and then project them into the domain of object recognition:
\begin{enumerate}
	\item What materials compose the system, either flowing through it or stationary?
	\item Which laws of nature describe their behavior?
	\item What are the external constraints of the system?
\end{enumerate}
If this article was about wind turbine design, the answers were evident. Air flows through the system, moving a turbine; the air flow is an external constraint, resulting from the weather; and the whole system obeys the laws of Newtonian physics and fluid dynamics.

The situation for imaging solutions appears to be quite the opposite: the designer is free to formulate any set of rules, and handle any subset of external input, as long as it works. Using the wind turbine analogy, the designer can control the laws of nature and even the weather to some extent.
While the physical approach to solution design will use a set of global rules defined outside its scope, the algorithmic one will define most of its rules internally and work out the adjustments for best results.  This has resulted in proliferation of algorithmic methods and solutions, each with its own internal reasoning.  I will use the term XMR (external model rules) and IMR (Internal model rules) for physical and algorithmic approaches respectively.

A good example of the difference between the two approaches is weather prediction: the algorithmic approach will look for a set of rules and processing steps or network architecture, while the physical approach will represent the atmospheric state by a set of equations and solve it numerically. They both estimate rain and temperature and both use statistics of past observations. However, IMR will use any set of algorithmic rules that leads to good results, while XMR will use a given set of physical rules and constrains to build a holistic physical model of the atmosphere. 

While IMR is about processing steps and their order, XMR sees the system as a non-hierarchical collection of elements, or even continuum, that are in balance, whether static or dynamic. XMR is about using the rules of nature to describe this balance. There is no algorithm or hierarchy behind fluid flow, therefore its calculation should rely the least possible on a specific set of algorithmic steps.

The reasons for using IMR for object detection and XMR for physical systems are evident: training a heuristic detection system is cheap, while none will look, for example, for optimal aircraft design solely by trial-and error. In addition, finding the imaging equivalent to questions 1-3 above is challenging; it requires formulating an \textbf{external } rule set that applies to any object detection case, regardless of specific case characteristics or designer preferences.

Object detection is generally IMR, with some XMR components, mainly statistic classifiers and regression methods. These two use externally defined rule sets, whether statistic or geometric, and can provide optimal generic solutions without significant case-specific algorithm adaptations. However, each one solves an isolated part of the detection flow.

A holistic XMR object detection framework can separate between the problem definition and its solution. Once the problem is well defined (using questions 1-3), the solution is merely a numeric issue, whereas for IMR, the solution itself tends to become most of the problem. 
 
\subsection{Structure of this document}
This document is structured as follows:
\begin{itemize}[leftmargin=*]
\item[--] Sec. \ref{Sec_XMR} represents {\OR} general approach with minimal details

\item[--] Sec. \ref{Sec_Relations} shows {\OR} relation to other methods. 

\item[--] Sec. \ref{Sec_OR_Root} provides a detailed description of {\OR}, including concepts and terminology (Sec. \ref{Sec_Concepts}) and some test cases (Sec. \ref{Sec_TestCases}), followed by a detailed probabilistic model (Sec. \ref{Sec_study}, \ref{Sec_DetectionPhase}).

\item[--] Sec. \ref{Sec_DemoCasesResults} presents test results for cases in Sec. \ref{Sec_TestCases}.

\item[--] Sec. \ref{Sec_implications} includes a discussion of {\OR} implications and the road ahead. 
\end{itemize}

\subsection{{\OR}: XMR object recognition framework}\label{Sec_XMR}
I will use the term \textit{ORS} for object recognition systems. Most ORS are composed of several processing steps, which may include filtering and grouping of internal results. For example,  face detection ORS could use local feature extraction followed by CNN on selected areas, and NMS to select the best match. Scientific works usually focus on the more interesting element (CNN in this case), and treat the rest as necessary "heuristic glue".
However, XMR ORS should model the whole process, and exclude only external constrains. This is necessary in order to model the output as direct result of the input, with minimal dependency on system internals.

ORS are built to receive evidence set $\ES = \{\ev_1, \ev_2, ... \ev_n \}$ and processes it to gain some knowledge about the object of interest.
ORS targets range from detecting object presence, to object localization, up to full model matching; each variant has many implementations depending on case-specifics. However, they all have one thing in common:
\textbf{all ORS  narrow down the range(s) of possible match(es) in the object model parameter space, by using externally-supplied evidence.} 

Model match algorithms operate in the model parameter space. For example, circle detection in 2D image will look for matches in $\{x, y, r\}$ space , where $x,y$ is the circle center and $r$ its radius. If $\MS$ is the model parameter space, $m$ is a point in $\MS$ representing a possible match, $\ES$ is the evidence set, and $\GM(\ES)$  is the probability density over $\MS$ that $\ES$ was generated by $m$:
\begin{equation}\label{GMN_intro_0}
	\GM(\ES) = p(m | \ES)
\end{equation}
than any ORS will search $\GM(\ES)$ or its marginals for significant ranges or maxima.
Full model match will seek a single salient range in $\GM(\ES)$; localization will attempt to narrow the range of  $\GM(\ES)$ spatial parameters; and object presence will try to verify that a salient range exists.

$\ES$ will typically contain both object-related evidence and random noise: $\ES = \ESC \cup \ESU$, 
where $\ESC$ is the subset of object-correlated evidence and $\ESU$ is the subset of random ones. Nearly all ORS will attempt to distinguish between them, either to meet design requirements (RANSAC model fitting for example) or to facilitate its operation. Any ORS that involves noise filtering, clustering or segmentation exercises some sort of evidence classification or grouping. These steps are usually treated as internal heuristics; however, they cannot be excluded from the XMR model. 

Following the above, $\GM(\ES)$ can be written as the weighted sum of all distributions resulting from the various $\{\ESC, \ESU\}$ grouping, each division assigned a weight according to its probability:
\begin{equation}\label{GMN_intro_1}
	\GM(\ES) = \sum_{i=1}^{ n_d} \bigg[P(\ESC_i, \ESU_i)\cdot p(m | \ESC_i, \ESU_i) \bigg]
\end{equation}
where $n_d = 2^{\ES}$ is the number of possible $\ES$ divisions into $\ESC$ (signal) and $\ESU$ (noise), $\ESC_i, \ESU_i$ are the sets resulting from division $i$, and $p(m | \ESC_i, \ESU_i)$ is $\m$ distribution given evidence sets \{$\ESC_i, \ESU_i$\}, where its weight $P(\ESC_i, \ESU_i)$ is the \textbf{probability} to get this division from all possible $n_d$ divisions. Note that as $\ESU_i$ are not object correlated, $p(m | \ESC_i, \ESU_i) \equiv p(m | \ESC_i) $. To make it clearer $P(\ESC_i, \ESU_i)$ can be written as:
\begin{equation}\label{GMN_intro_2}
\begin{split}
	& P(\ESC_i, \ESU_i) = \frac{Q(\ESC_i, \ESU_i)}
	{\sum_{k=1}^{ n_d}Q(\ESC_k, \ESU_k)}\\
&\text{ where }\\
	& Q(\ESC, \ESU) = \int\limits_{\MS} p(\ESC, \ESU | \m) \cdot p(\m) \diff\m
\end{split}
\end{equation}
Expansion of Eq. \ref{GMN_intro_1} is further developed Sec. \ref{Sec_DetectionPhase}, Eq. \ref{EqDet_Division_1}.

Most ORS involve a combination of several tasks; they all operate in the same object model parameter space, each focusing on a limited range of space or parameters.
However, ORS do not attempt to calculate $\GM(\ES)$ explicitly. 
{\OR}, on the other hand, was designed to calculate $\GM(\ES)$, and to continuously update it with incoming evidence. It was \textbf{not} designed to solve a detection or regression problem, but rather enables deriving them from $\GM(\ES)$ when needed. Note that:
\begin{enumerate}
	\item Both $\GM(\ES)$ and $\ES$ are implementation-independent, resulting only from the statistics and geometry of $M$. 
	\item $\GM(\ES)$ contains all knowledge about possible matches, not only the best match or the distance from it. It can encapsulate ambiguous situations like multi-modality or lack of information.
	\item $\ES$ is a random collection of random variables of various types, and can be of any size.
	\item  $\GM(\ES)$ is invariant to the order of $\ES$  member; any permutation of $\ES$ should yield the same result.
	\item Eq. \ref{GMN_intro_1} uses the same probability space and distribution for various $\ES$ divisions; hence actions like evidence grouping, noise removal and outliers detection can be an integral part of an ORS using this model.
\end{enumerate} 
Probability densities are usually constructed by collecting observations; 
doing so for $\GM(\ES)$ directly is practically impossible, as its random variable $\ES$ can contain any number of members, of various types. Even limiting it to some large number and discarding permutations will require collecting huge amounts of observations, and constructing a separate $\GM(\ES)$ for each type combination. 

{\OR} solves this by using  $\GM(\ev \in \ES)$ instead. Every single evidence $\ev$ in $\ES$ holds some information about possible matches in $\MS$. A corner of an object, for example, narrows the match ranges considerably; edge elements have a lesser narrowing effect, but even a color patch has some information about the ``whereabouts'' of possible matches. Put more formally, this information is expressed as $\GM(\ev)$, the probability distribution over $\MS$  that evidence $\ev$ was generated by model parameters $m$ :
\begin{equation*}
	\GM(\ev) =  p(m|\ev)
\end{equation*}
{\OR} uses  $\GM(\ev)$ to calculate  $\GM(\ES)$ as a stationary Markov process  where a new evidence is added to $\ES$ at each step.
The process state at step n+1, $\GM(\ES_{n+1})$, can be calculated as the joint distribution of $\GM(\ES_n)$  and	$\GM(\ev_{n+1})$ :
\begin{equation}\label{intro_1}
	\GM(\ES_{n+1}) = J(\GM(\ES_{n}), \GM(\ev_{n+1}))
\end{equation}
where $J$ is an operator that creates the joint distribution over $\MS$ of $\GM(\ES)$ and a single new evidence $\ev$; this operator is developed in Sec. \ref{Sec_DetectionPhase}

Implementation of Eq. \ref{intro_1} requires developing operator $J$, and building $\GM(\ev)$. While the first will be valid for all models, the second depends both on the specific model statistics and evidence type, and requires per-model data gathering and analysis. In this work I have created $\GM(\ev)$ for each model and evidence type by generating synthetic evidence and fitting a GMM (Gaussian mixture model) to their populations.

$\GM(\ES)$ is implemented using GMM (Gaussian mixture model); detection, regression, and any other variants are implemented as operations on it. For example, if  $\GM(\ES)$ is close enough to uni-modality with small variance, then its peak describes an exact match. Note that "close enough to uni-modality with small variance" implies having several case-specific parameters; {\OR} makes a distinction between the process of building and maintaining $\GM(\ES)$, which is nearly clean of any heuristics, and the case-specific operations on it, which requires external tolerance parameters. 

Therefore, {\OR}'s most important characteristics are:\textit{ `always keep all options on the table'} - maintain a continuous distribution of possible matches, and update it on each new evidence. In addition, process stationarity implies that the state should not depend on evidence order; any permutation of an evidence set should yield the same state. 
These  characteristics are, more or less, the \textbf{antithesis} of conventional algorithm design; algorithms have an ending point where the results are generated, they prefer to narrow down the possible solution range and reduce dimensionality wherever possible, they usually depend on strictly structured input, and they process it in a predefined order of actions.

The idea of a process state that is valid all along the continuous flow of incoming evidence may appear unconventional. However,consider a human driver accessing a crossroad: information flows in continuously, and is immediately used to update the world model. This information is used to prioritize actions and acquisition of additional data (where to look, what to check). Similarly, valid $\GM(\ES)$ can be used for optimal resource usage.
Valid $\GM(\ES)$ has value far beyond autonomous cars or robotics; as $\GM(\ES)$ has a concrete, absolute probabilistic meaning, it can be the basis of an \textbf{open architecture} where several such processes update and use each other's state. This is not possible with algorithms like NN, where the state has internal meaning only.

\subsubsection{ORCEA innovative implications}
\begin{enumerate}
	\item \textbf{Explainable, transferable model:} 
	As described in Sec. \ref{Sec_DetectionPhase}, $\GM(\ES)$ captures the entire solution space and represents it as a set of concrete probabilistic hypotheses about the object, which can be used across algorithm boundaries.
	On the importance of explainable AI see \citet{Conc_AeplanatoryAI}. 

\item \textbf{Unified framework for the entire detection process:}
	As \citet{Conc_3R} have stated, detection is implemented usually as "the three R's" - recognition, reconstruction and reorganization, three separate processes which can benefit from interacting with each other. {\OR} supplies such a framework, as regression, noise exclusion, and classification are all integral parts of the same probabilistic model. 
	
	\item \textbf{Minimal algorithm and heuristics:}
	The detection `algorithm' is simple and uniform for all cases; it consists basically of building and updating a joint distribution, based on incoming events.
	As $\GM(\ES)$ and  $\GM(\ev)$ are both represented by GMM, and there is no use of heat-maps or any other volume discrete mapping, the cost of dimensionality is low, and operations such as calculating marginal or conditional distributions are trivial.
	The order of event processing is irrelevant, except for numerical reasons, as it consists of distribution multiplications.
	{Any type of evidence can be integrated:} no heuristics are needed, as they are all expressed in the same $\MS$.

	\item \textbf{Updating and extending existing model with new data is trivial:} {\OR} has no heuristic parameters, it solely requires \GMevm, which is calculated directly from observations and can be easily updated with new ones. For example, \GMevm can be calculated as a weighted average with higher weight for latest data, or as sliding window average over time, to enable smooth transition from synthetic objects to real-world ones. Model dimensionality can be extended easily as well, for example in the case of newly discovered latent variables, by assigning an a-priory distribution to the new variables, and updating it gradually with incoming observations.

\end{enumerate}

\subsection{Relation to other methods}\label{Sec_Relations}
{\OR} cannot yet be compared by benchmarks; in this section I compare its design principles with those of other object recognition methods. I focus on their solution space: Is it $\MS$? How do they map it? 

\paragraph{Generalized Hough Transform (GHT):}
The starting point of GHT (\citet{GHT_Ballard}) has much in common with \OR. They both operate in the model parameter space $\MS$, and both project a set of evidence $\ES$ directly on $\MS$ to create a mapping of $\MS$. However, instead of constructing $\GM(\ES)$, GHT creates an `R-table', which is a discrete table over $\MS$ that accumulates the projections of each evidence. The contribution of each evidence is not expressed as the PDF \GMevm, but as a discrete distribution of weights over a small area in $\MS$. The best match is the cell with maximum accumulated contributions. This raises several issues:
\begin{itemize}
	\item The sum of evidence contributions is not equivalent to their joint distribution, therefore the R-table is not equivalent to $\GM(\ES)$. The R-table has no concrete statistical meaning; at best, it is a monotonically ascending function of $\GM(\ES)$, without concrete meaning outside its scope.
	\item The resources required for discrete mapping of $\MS$ and operations on it grow exponentially with dimensionality.
	\item GHT uses edge elements only. Theoretically it can support area patches or any other object-related evidence; but their $\GM(\ev)$ equivalent of discrete weights might be too spread or difficult to calculate.
\end{itemize}
As a result, most of the works related to GHT concentrate on optimizing the process of evidence selection and projection (\citep{GHT_Survey, GHT_PHT, GHT_RHT}). 

\paragraph{Regression, RANSAC \citep{RANSAC_Fischler1981RandomSC}:}
Given evidence set $\ES$, regression methods will look for point $\m_{opt}$ in $\MS$ that minimizes a distance function $\Delta_m(\ES, \m)$ between $\ES$ and the object resulting from $\m.$ The distance function will typically use spatial mismatch, but can include other differences.
This process requires several iterations starting from a reasonable initial guess. At each iteration outliers are detected using RANSAC (\citet{RANSAC_Fischler1981RandomSC}), \mm is refined accordingly, and  $\Delta_m(\ES, \m)$ is recalculated. As these steps are resource-expensive, regression methods will not try to construct $\GM(\ES)$ or cover $\MS$, but to optimize the search for $\m_{opt}$. 

Let's look at two examples of 3D cylinder detection using regression: \textbf{\citet{REG_Cylinder1}}
detect a cylinder in a 3D point-cloud. The initial guess is created using local neighborhood properties (normal vectors and curvature); then follows an iterative process of cylinder axis vector approximation, points projection on a plane normal to this vector, 2D circle fitting, and outliers rejection. Finally, the cylinder endpoints are calculated from inliers. 
\textbf{\citet{REG_Cylinder2}} choose another path: they use ROBPCA (\citet{REG_Cylinder2_RPCA}) to detect cylinder axis without outliers, then follow as above, by inliers projection and circle fit. Their solution looks iteration-less, but the costly iterations are encapsulated in ROBPCA.

Note that the main effort in these works is dimensionality reduction: separately detect the axis direction, the radius, and the endpoints. This is typical to many detection algorithms that rely on non-trivial regression; the distance function $\Delta_m(\ES, \m)$ might be simple, but convergence requires a good first guess, usually by searching is lower-dimension space.

\paragraph{Statistic classifiers:}
Unlike {\OR}, statistic classifiers do not operate in $\MS$, but each classifier operates in an object feature space according to its design. In addition, the input cannot be a general set of evidence but has to be cast into a feature vector with predefined structure. Statistic classifiers are the right tool for calculating the probability of an object to belong to one of several classes, given a feature vector; however, in the case of object recognition they are largely used as a regression tool, hence solving a different problem than the one they were designed for. \citet{CLA_Viola} seminal article on face detection supplies a good example: a set of classifiers were trained to distinguish between windows containing a centered face and those without a face. In the detection phase, a sliding window was used to classify patches accordingly. \citet{CLA_Li_SURF_face_car} applied a similar method for multi-view face detection using SURF descriptors, where a classifier for each view (front, half profile, profile) was trained separately. This raises the question: Is $\sfrac{1}{2}$ a face considered a hit? What about $\sfrac{3}{4}$? In both works the authors solved it by an additional heuristic phase of local maxima selection. The point is that the object (face) is defined in the image space or \{image , view-angle\} space, but the classifiers operate in their feature space (SURF or Haar-like descriptors in those cases), thus used like a regression distance function on $\MS$.

\paragraph{Artificial Neural Networks:}
The root difference between ANN and {\OR} is that
ANN are black box by design, built to `grow' their set of rules internally, while {\OR} is a knowledge-based white box, where model internals have global probabilistic meaning. As described in Sec. \ref{Sec_DetectionPhase}, both $\GM(\ES)$ and \GMevm represent concrete probabilistic hypotheses about the object.

I am sometimes asked if {\OR} can be implemented using ANN. 
Theoretically it can be done, as ANN is a generic mapping tool that can perform any input-to-output mapping, including  $\ES$ to $\GM(\ES)$; but it will require a manually prepared training set of pairs $\{\ES , \GM(\ES)\}$. ANN handles masses of labeled data well, but defining $\GM(\ES)$ manually for huge amounts of $\ES$ variants is impossible.

There are several works of interest in this context. \citet{ANN_RegHM_Bulat_2016} article on pose estimation proposes a detection-followed-by-regression CNN cascade, where the first part creates one heatmap per each body part indicating where the part is likely to be, 
and the second part performs regression on these heatmaps.
In terms of {\OR}, each of these 2D heatmaps is a discrete function $F$ of the marginal distribution of  $\GM(\ES)$ for a single body part; as with GHT, $F$ is \textbf{not} PDF, but at best a monotonically increasing function of it.
\citet{ANN_RegDF_Toshev_2014} take a different approach to the same problem: they train a network to detect directly the pose vector, which defines the location of all body parts. The point is that in both cases the labeled data consists of a set of locations and body parts; no matter what network architecture one will use, this training data cannot magically create $\GM(\ES)$, it can at best find its maximum, or create heatmaps around it.

\section{{\OR} Model}\label{Sec_OR_Root}
\subsection{Concepts and terminology}\label{Sec_Concepts}
\paragraph{Models, instances and the model parameter space:}
A typical object in our case can be described by a generic model $M$ (circle, line...) and a set of model parameters  $m$ (diameter, length...). Some of the parameters are structural, and some are color or pattern related. $M$ parameter-space is $\MS$.
For example, if $M$ is an upright rectangle with specific edge width, as in Sec. \ref{DemoCase1}, then it is defined by its center $\{x , y\}$, width $w$, height $h$, and border thickness $b$, and the model parameter space is:
\begin{equation*}
	\MS = \{x, y, w, h, b\}
\end{equation*}
A specific instance is represented by point $m$ in $\MS$. The a-priory probability density of $m$ is $\GAM$ over $\MS$.

\paragraph{Evidence - Edge and area elements:}
The input to any object recognition process is a set of evidence $\ES$, of various types: edge elements, local descriptors, line segments, curve segments, color patches etc. The evidence used in this work are edge and area elements; \textbf{they were not detected in images, but created programatically from each model's parameters} as explained in section \ref{Sec_DemoCasesResults}. Two types of evidence are created:
\begin{itemize}
	\item \textbf{EE:} Edge element, representing local maximum response to some edge kernel. Its spatial properties are location $x,y$ and orientation $\theta$, and shape-related ones are edge width $w$ and amplitude $a$. EE parameter space is $\Phi_e = \{x , y, \theta, w, a\}$
	
	\item \textbf{AE:} Area element representing small rectangular areas of uniform color or texture of interest. For example, shades of green for apple detection, black and white for checker-board like target detection. AE parameters are center  $x,y$ and size $w$, and a value $c$ representing its color or texture. AE parameter space is $\Phi_a = \{x , y, w, c\}$
\end{itemize}

\paragraph{Error and noise:}
There are several phenomena that degrade evidence sets:
\begin{enumerate}
	\item \textbf{Missing coverage:} Edge parts that are not represented by EE, or object parts with the color/texture of interest not represented by AE, due to occlusion, object variability, poor image quality or problematic preprocessing.   
	\item \textbf{Noise:} Random evidence, not correlated with the object of interest. For example random strokes (line segments, curves) creating edge elements and random blobs creating area elements.
	\item \textbf{Scatter:} Small deviations of evidence location and properties, resulting from measurement or calculation inaccuracies, or object variability.
\end{enumerate}
The term \textbf{evidence set quality} refers to the level of missing coverage, noise and scatter in the evidence set.

\subsection{Test cases}\label{Sec_TestCases}
Three test cases are used to demonstrate and test {\OR}, described in Sec. \ref{DemoCase1} - \ref{DemoCaseLast}: upright rectangle, 4X4 grid, and spiral sector. 

\begin{figure}[H]
	\centering
	\begin{subfigure}[t]{0.27\linewidth}
		\includegraphics[width=\linewidth]{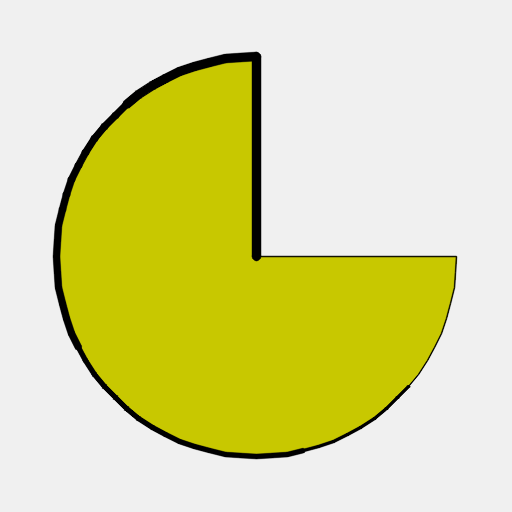}
		\caption{Theoretical object}
	\end{subfigure}
	\begin{subfigure}[t]{0.27\linewidth}
		\includegraphics[width=\linewidth]{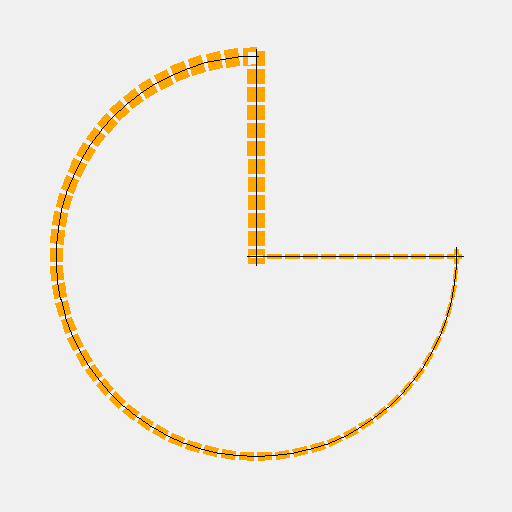}
		\caption{Ideal EE set}
	\end{subfigure}
	\begin{subfigure}[t]{0.27\linewidth}
		\includegraphics[width=\linewidth]{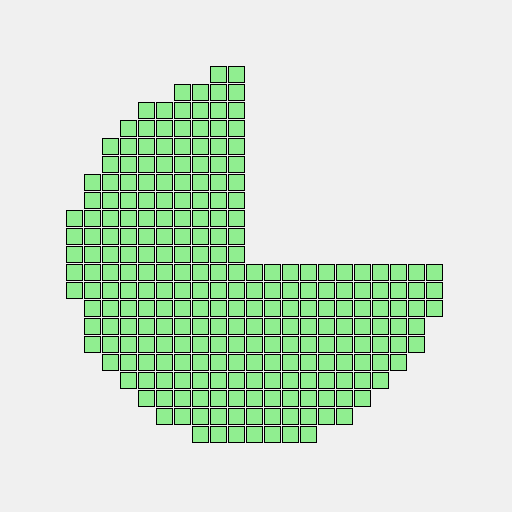}
		\caption{Ideal AE set}
	\end{subfigure}
	\begin{subfigure}[t]{0.27\linewidth}
		\includegraphics[width=\linewidth]{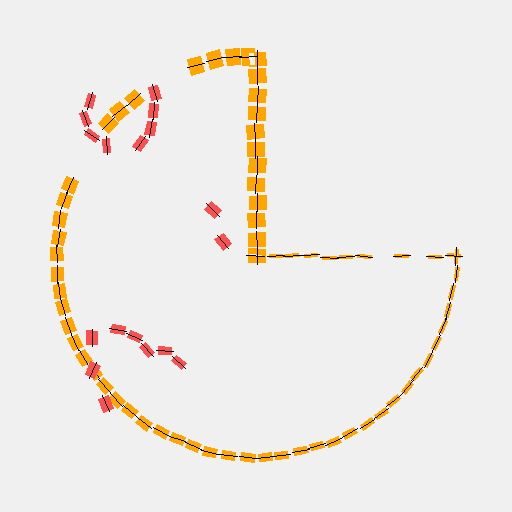}
		\caption{EE, high quality}
	\end{subfigure}
	\begin{subfigure}[t]{0.27\linewidth}
		\includegraphics[width=\linewidth]{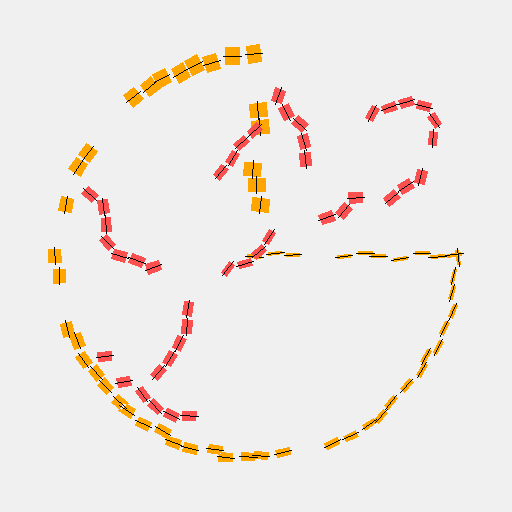}
		\caption{EE, med. quality}
	\end{subfigure}
	\begin{subfigure}[t]{0.27\linewidth}
		\includegraphics[width=\linewidth]{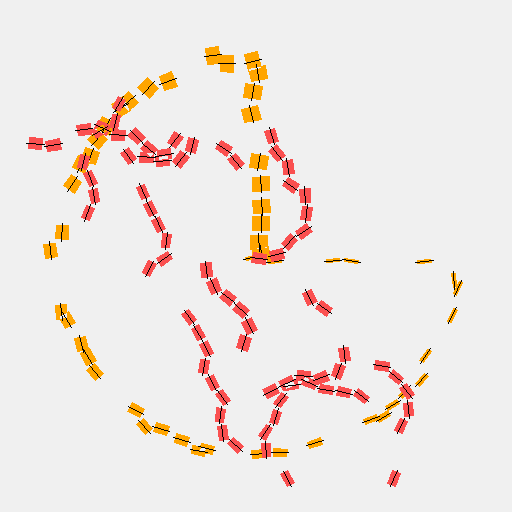}
		\caption{EE, poor quality}
	\end{subfigure}
	\begin{subfigure}[t]{0.27\linewidth}
		\includegraphics[width=\linewidth]{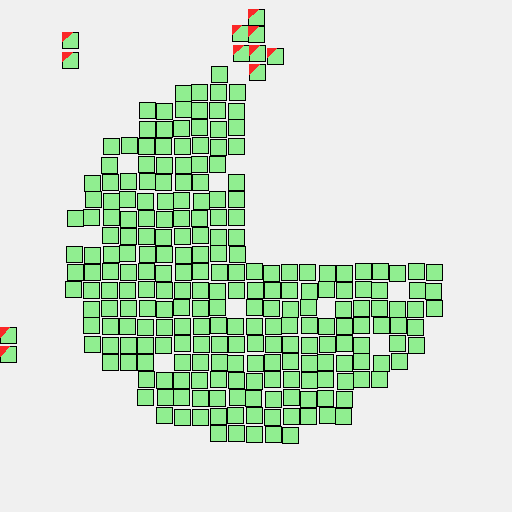}
		\caption{AE, high quality}
	\end{subfigure}
	\begin{subfigure}[t]{0.27\linewidth}
		\includegraphics[width=\linewidth]{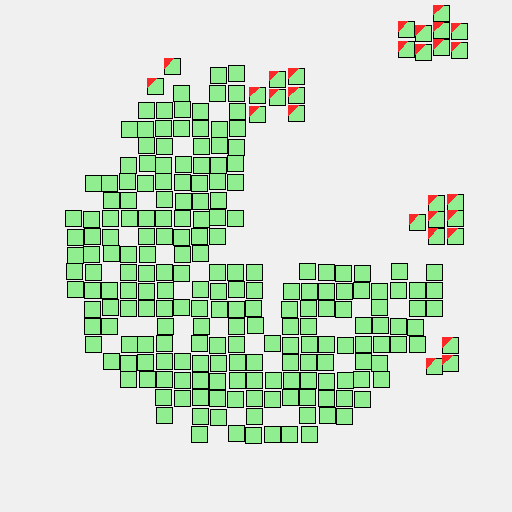}
		\caption{AE, med. quality}
	\end{subfigure}
	\begin{subfigure}[t]{0.27\linewidth}
		\includegraphics[width=\linewidth]{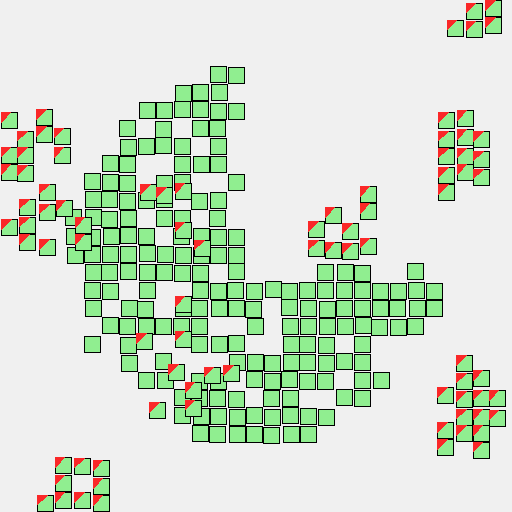}
		\caption{AE, poor quality}
	\end{subfigure}
	\caption{Case visualization}
	\label{fig:DataVisualization}
\end{figure}

\paragraph{Data Visualization:}
The simulated data has to be visualized for report and analysis, here and in  \href{http://www.odedcohen.com/orex/v2/index.html\#test\_results}{\textsf{http://www.odedcohen.com/orex/v2/index.html\#test\_results}}.
The following  visualization types are used: 
\begin{itemize}
	\item\textbf{ Theoretical object}, showing the shape without any degradation; see Fig. \ref{fig:DataVisualization}.a. The edge width is shown, and the yellow filling represents any detectable color or pattern.	
	\item \textbf{EE set}; see Fig. \ref{fig:DataVisualization}.b. The black line shows the edge center, and the color rectangle around is shows the EE length and width (edge thickness). 
	Fig. \ref{fig:DataVisualization}.d - \ref{fig:DataVisualization}.f show 3 levels of EE quality, with scatter and missing coverage. Random noise was added as well, shown in red.
	\item\textbf{ AE set}, showing small areas (patches) where some color or pattern was detected; see Fig. \ref{fig:DataVisualization}.c. 
	Fig. \ref{fig:DataVisualization}.g - \ref{fig:DataVisualization}.i show 3 levels of AE quality, with scatter and missing coverage. Random noise was added as well, represented by elements with red in the upper-left corner. In cases where there are several classes of AE, the various classes are represented by different colors.
\end{itemize}

\pagebreak
\subsubsection{Test case 1: Upright rectangle}\label{DemoCase1}
This object is defined by its center $\{x , y\}$, width $w$, height $h$, and border thickness $b$, hence the model parameter space is:
\begin{equation*}
	\MS = \{x, y, w, h, b\}
\end{equation*}
\begin{figure}[H]
	\centering
	\begin{subfigure}[b]{0.35\linewidth}
		\includegraphics[width=\linewidth]{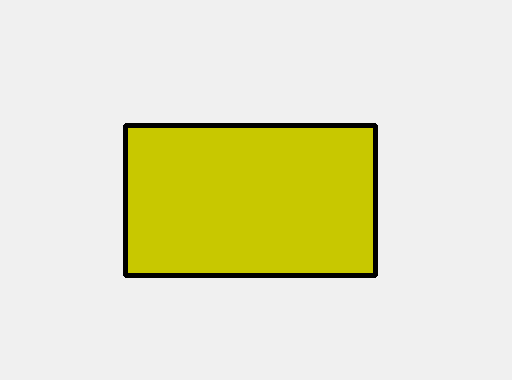}
		\caption{$w$=150, $h$=100, $T_B$=9}
	\end{subfigure}
	\begin{subfigure}[b]{0.27\linewidth}
		\includegraphics[width=\linewidth]{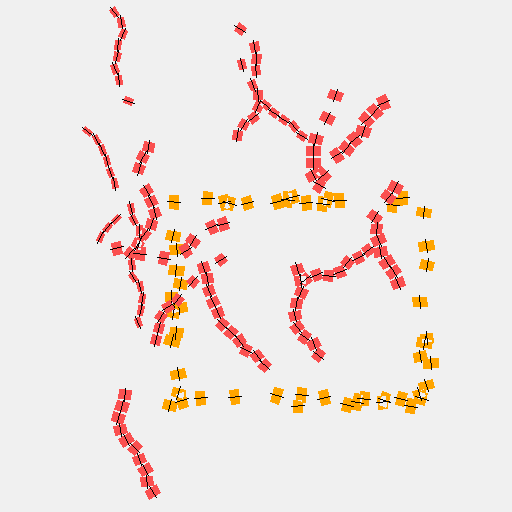}
		\caption{Low quality}
	\end{subfigure}
	\begin{subfigure}[b]{0.27\linewidth}
		\includegraphics[width=\linewidth]{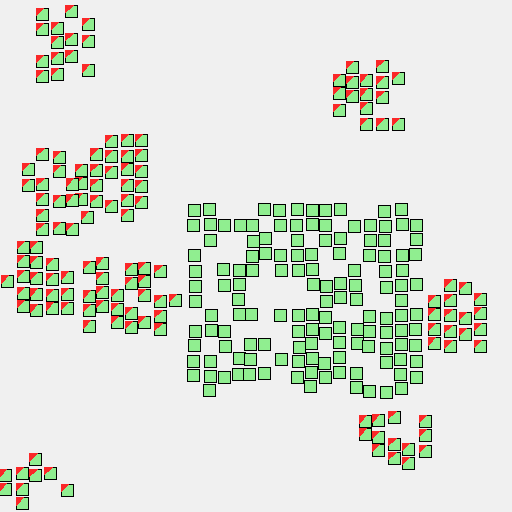}
		\caption{Low quality}
	\end{subfigure}
	\caption{Upright rectangle.}
	\label{fig:UprightRectExamples}
\end{figure}

\subsubsection{Test case 2: 4X4 grid}\label{DemoCase2}
This is a square 4X4 grid with alternating color, similar to checkerboard or calibration target. It is defined by its center $x,y$, width $w$, angle $\theta$, and border thickness $b$, hence the model parameter space is:
\begin{equation*}
	\MS = \{x, y, w, \theta, b\}.
\end{equation*}
Every detected AE has also a `color' class which corresponds to one of the colors of the grid; some AE were assigned the wrong color class. As AE quality decreases, more AE class miss-assignments occur.

\begin{figure}[H]
	\centering
	\begin{subfigure}[b]{0.3\linewidth}
		\includegraphics[width=\linewidth]{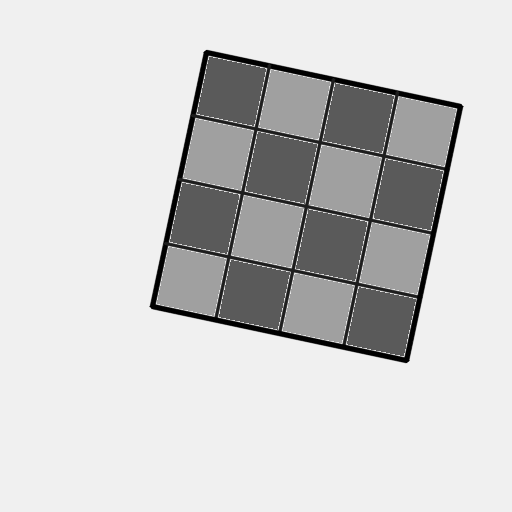}
		\caption{$w$=250, $\theta$=30$^\circ$}
	\end{subfigure}
	\begin{subfigure}[b]{0.3\linewidth}
		\includegraphics[width=\linewidth]{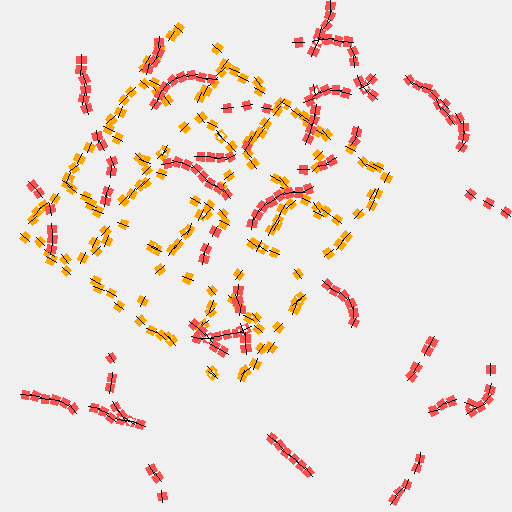}
		\caption{Edge elements}
	\end{subfigure}
	\begin{subfigure}[b]{0.3\linewidth}
		\includegraphics[width=\linewidth]{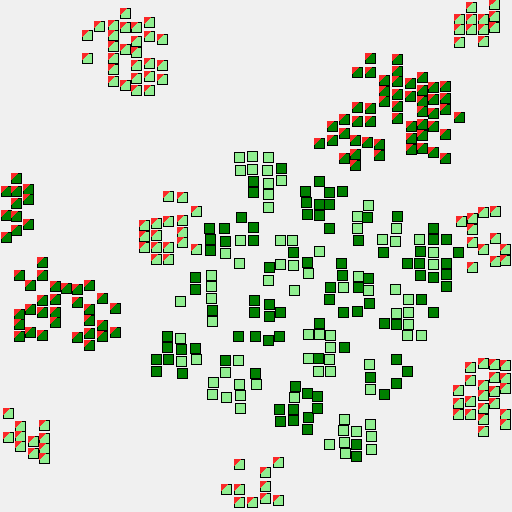}
		\caption{Area elements}
	\end{subfigure}
	\caption{4X4 grid.}
	\label{fig:GridExamples}
\end{figure}

\subsubsection{Test case 3: Spiral sector}\label{DemoCaseLast}
This object is a sector of logarithmic spiral, defined by its center, orientation $\theta_0$, starting radius $R_0$, angular span $\Theta$, and growth exponent $b$.
The polar coordinates relative to its center are given by:
\begin{equation}\label{eqLogSpiral}
	r(\alpha) = R_0 \cdot e^{b(\alpha- \theta_0)} \:;\: \theta_0 <= \alpha <= \Theta + \theta_0
\end{equation}
It forms a closed contour by connecting the end points to the center. In addition, its area has some typical pattern or color.
\begin{figure}[H]
	\centering
	\begin{subfigure}[t]{0.24\linewidth}
	\end{subfigure}
	\begin{subfigure}[t]{0.24\linewidth}
		\includegraphics[width=\linewidth]{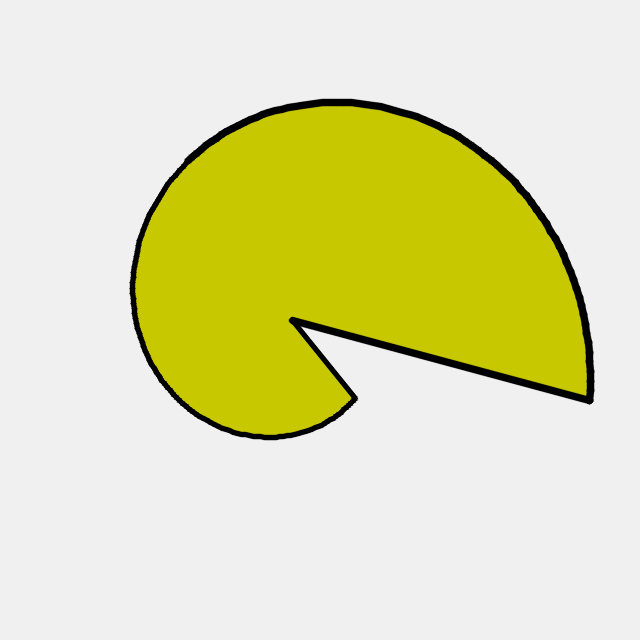}
		\caption{$b=2, \Theta=1.2\pi $}
	\end{subfigure}
	\begin{subfigure}[t]{0.24\linewidth}
		\includegraphics[width=\linewidth]{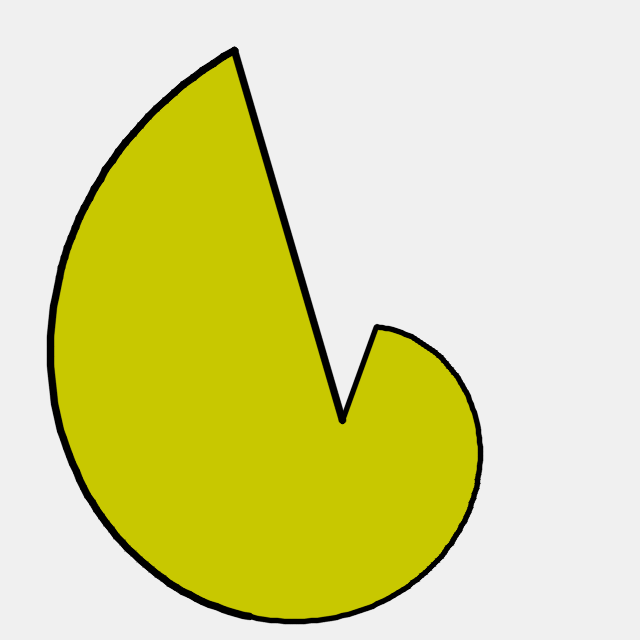}
		\caption{$b=1, \Theta=0.1\pi $}
	\end{subfigure}
	\begin{subfigure}[t]{0.24\linewidth}
		\includegraphics[width=\linewidth]{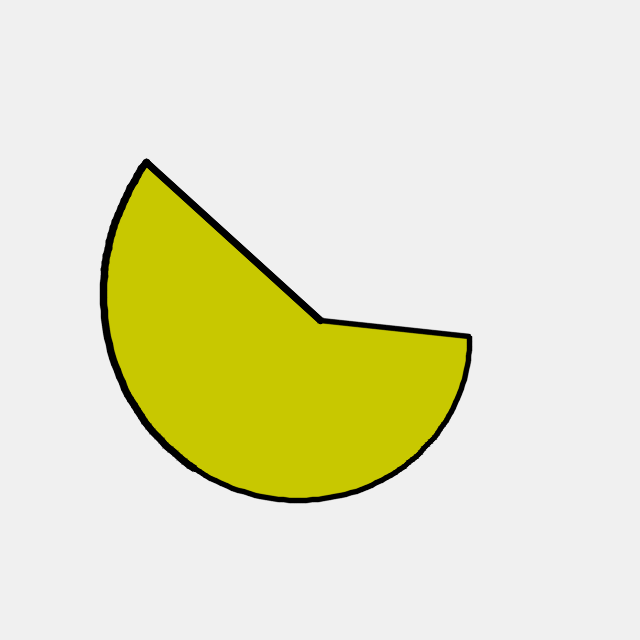}
		\caption{$b=1, \Theta=0.1\pi $}
	\end{subfigure}
	\begin{subfigure}[t]{0.24\linewidth}
		\includegraphics[width=\linewidth]{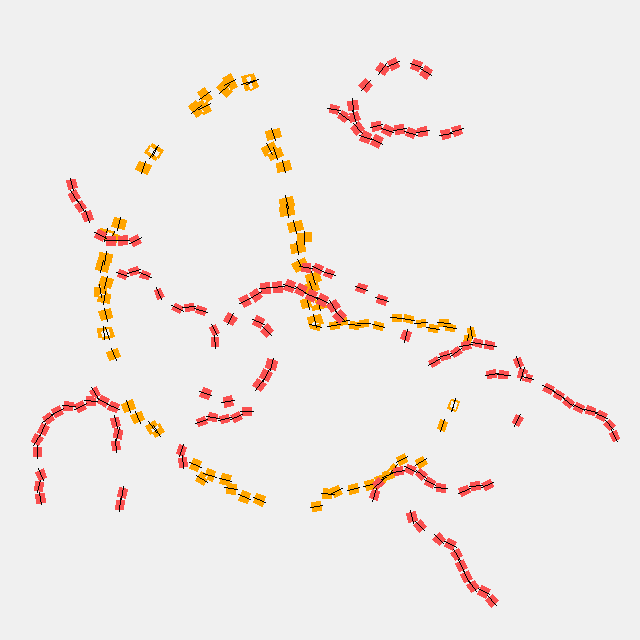}
		\caption{Low quality}
	\end{subfigure}
	\caption{Spiral sector examples.}
	\label{fig:coffee}
\end{figure}

\subsection{Study phase} \label{Sec_study}
{\OR} is a generic framework, totally ignorant of specific cases; all model-specific information is encapsulated in \GMevm, which connects between evidence and generating instances. 
Any evidence $\ev$ in $\ES$, either EE, AE or other, has information that can further narrow $\GM$. This information can be expressed as the \textbf{probability density that $\ev$ was generated by an instance $m$ in $\MS$}:
\begin{equation}
	\GMev = p(\m|\ev)
\end{equation}

\GMevm can be seen as a map in $\MS$ indicating which combinations of parameters are more likely to create $\ev$. 
In most cases this information is too scattered to draw any decisive conclusion about the object of interest based solely on $\ev$; for example, in the grid case (Sec. \ref{DemoCase2}), an EE can be part of any of the 10 grid lines, or just random noise.
There aren't many ``silver bullets'', which narrow  \GMevm dramatically (for example, a corner evidence when detecting rectangles), and {\OR} does not depend on them. 

An \textbf{observation} $\ob(\ev, m)$ is the combination of evidence $\ev$ and the parameters $m$ of the instance that created it.
An observation requires knowledge of $m$, therefore it can be created only during supervised learning.
For example, if $m$ is an upright rectangle (Sec. \ref{DemoCase1}) and \evm is an EE, then $\ob$ will consist of the EE features plus the point $m$ in $\Phi_{\mathsf{M}}$:
\begin{equation}
	\ob(\ev, m) = \{\overbrace{x_e, y_e, \theta_e , w_e, a_e}
	^{\text{{\large $\ev$}}} 
	\; , \;
	\overbrace{x_m, y_m, w_m, l_m, b_m}
	^{\text{\large $\m$}} \} 
\end{equation}
Given enough observations, it is possible to build their distribution $\GOBo$, and derive \GMevm from it:
\begin{equation}
	\GOBo = p(m, \ev)
\end{equation}
\begin{equation}
	\GMev = p(m | \ev) = p(m, \ev) / p(\ev)
\end{equation}
\GMevm is built by collecting observations and fitting a GMM to the population. Observations can be collected from real-world samples, synthetic cases, or both. In very simple cases \GMevm might be calculated directly.
In this work, observations were created programatically: for each model a large number of instances was created to cover a predefined range of parameters, and noise, scatter and missing coverage were added. Using real-world input will be explored in future works; anyway, even then the first version of \GMevm will probably be synthetic to facilitate the training.
	
Approximating \GMevm by GMM is not straight-forward. 
\GMevm is not a classic candidate for GMM approximation, as it was not created from different populations; approximating it by GMM resembles approximating a smooth curve by a set of lines. Therefore the number of Gaussian components is simply a compromise between resolution and performance.

\subsection{Detection phase}\label{Sec_DetectionPhase}
\subsubsection{Probability spaces}
Before going into the detection phase details, it is necessary to define the various probability spaces that are involved, using the previous definitions and some new ones:
\paragraph{Model probability space} = $( \MS , p(\m))$ where $\MS$ is the model space, \mm is a point in $\MS$, and $p(\m)$ is the probability density of \mm over $\MS$. Detection final result is expressed as match range(s) in $\MS$. The apriori distribution of $\m$ is $\GAM$.

\paragraph{Evidence probability space} $ = ( \evS , p(\ev))$ where \evm is a single evidence, and $\evS$ is the evidence space for this evidence type.
As there are various evidence types, I will use superscript where necessary: $\evSe$ for edge element, $\evSa$ for area element. 
The probability density of \evm over $\evS$ is $p(\ev) = p_c(\ev) + p_u(\ev)$ where $p_c(\ev)$ and $p_u(\ev)$ are the distribution parts of object-correlated evidence and uncorrelated noise respectively.

\paragraph{Evidence-set probability space} $ = ( \ESS , p(\ES))$ where $\ES$ is a set of evidence which might be of various types, $\ESS$ is the product of their evidence spaces:
\begin{equation}
\ESS = \evS_1 \times \evS_2... \times \evS_n = \prod_{E} \evS
\end{equation}
and $p(\ES)$ is probability density of \textbf{the set of evidence} $\ES$ over $\ESS$. $\ES$ will typically contain both object-correlated evidence and uncorrelated noise: $\ES = \ESC \cup \ESU$; the corresponding probability spaces will be 
$(\ESCS , p_c(\ESC))$ and $(\ESUS , p_u(\ESU))$.

%\paragraph{Observation probability space:} Observations tie together evidence and model. For a single observation $\ob$ the probability space is $( \obS , p(\ob))$, where 
%$\ob = (\ev, \m)$, and  $\obS = \MS \times \evS$.
%
%\paragraph{Observation-set probability space:} An observation set is the conclusive collection of  observations about the object. The corresponding probability space is $( \OBSS , p(\OBS))$, where 
%$\OBS = (\ESS, \m)$, and  $\OBSS = \MS \times \ESS$.
%
%\paragraph{Putting it all together:} Our accumulated experience regarding evidence and model M creates the observation-set space. When M is marginalized out, we have the evidence space. Using the conditional distribution of $\ESS | m$, we can reconstruct $\m$

\subsubsection{Calculating and updating $\GM(\ES)$ with incoming evidence}
In the following section I added the space symbol they are related to under probability distributions, where I thought it might add clarity. For example, $\ESp(\ES)$ instead of just $p(\ES)$.

According to Eq. \ref{GMN_intro_1} and  \ref{GMN_intro_2}, $\GM(\ES)$ can be written as:
\begin{equation}\label{EqDet_Division_1}
	\GM(\ES) = const\cdot\sum_{k=1}^{ n_d} \bigg[\ESp(\ESC_k, \ESU_k)\cdot \MSp(m | \ESC_k) \bigg]
	= const\cdot\sum_{k=1}^{ n_d} \bigg[\ESUp(\ESU_k)\cdot \ESCp(\ESC_k)\cdot \MSp(m | \ESC_k) \bigg]
\end{equation}
where $n_d$ is the number of possible $\ES$ divisions into  $\ESC$ (object-correlated evidence) and $\ESU$ (noise, uncorrelated evidence),  $\ESC_k$ and $\ESU_k $ are the sets resulting from division $k$, and $p(\ESC_k, \ESU_k)$ is the probability density over $\ESS$ to get division $k$.  $\GM(\ES)$ is a weighted sum of distributions resulting from possible divisions of $\ES$. 

Assuming noise evidence are uncorrelated:
\begin{equation}
	\ESUp(\ESU) = \prod_{u \in \ESU}p(u)
\end{equation}
$\ESC$ members cannot be assumed independent in general; however as they depend only on $m$, we can assume independence for a given $m$
\begin{equation}
	p(c_i, c_j | m) = p(c_i|m)\cdot p(c_j|m)
\end{equation}
Lets define the term inside the summation in Eq. \ref{EqDet_Division_1}  as:
\begin{equation*}
		\GM(\ESC, \ESU) = \bigg[\ESUp(\ESU)\cdot \ESCp(\ESC)\cdot \MSp(m | \ESC) \bigg]
\end{equation*}
Then
\begin{equation}
	\begin{split}
\GM(\ESC, \ESU) & = \bigg[\prod_{u \in \ESU}p(u)\bigg] \cdot p(\ESC)\cdot \bigg[\frac{p(\ESC | \m) \cdot p(\m)}{p(\ESC)}  \bigg] \\
 & = \bigg[\prod_{u \in \ESU}p(u)\bigg] \cdot \bigg[p(\m) \prod_{c \in \ESC} p(c | \m)  \bigg] \\
 & = \bigg[\prod_{u \in \ESU}p(u)\bigg] \cdot \bigg[p(\m) \prod_{c \in \ESC} \frac{p(\m | c) \cdot p(c)}{p(\m)}  \bigg] \\
& = \GAM \cdot K_{\ESU} \cdot K_{\ESC} \cdot \Gamma_m(\ESC)
	\end{split}
\end{equation}
where
\begin{align*}
	\GAM & = p(\m), \text{ the a-priori \mm distribution }\\
	\Gamma_m(\ESC) & = \prod_{c \in\ESC} \Gamma_m(c), \text{ where }
	\Gamma_m(c)  = \frac{p(\m | c)}{\GAM} \\
	K_{\ESC} & =  \prod_{c \in \ESC} p_c(c)\\
	K_{\ESU} & =  \prod_{u \in \ESU} p_u(u)
\end{align*}

\paragraph{$\GM(\ES)$ can now be formulated as a Markov process.}
During detection, a new evidence \evm is added to $\ES$ at each step;
every existing division - $\{\ESC ,\ESU\}$  is extended to two new divisions - $\{\ESC+\ev, \ESU\}$ and $\{\ESC, \ESU+\ev\}$ reflecting the probability that \evm is object-related or noise accordingly. 
If \evm is noise, only $K_{\ESU}$ has to be updated:
\begin{equation}
	K_{\ESU}(\ESU+ \ev) = K_{\ESU}(\ESU)\cdot p_u(\ev)
\end{equation}
If \evm is object-related, $K_{\ESC}$ and the distribution itself have to be updated
\begin{align}
	K_{\ESC}(\ESC+ \ev) & = K_{\ESC}(\ESC)\cdot p_c(\ev) \\
	\Gamma_m(\ESC+ \ev)  & = \Gamma_m(\ESC)\cdot \Gamma_m(\ev)
\end{align}
If $\ES_{n}$ is the evidence set at step $n$, and evidence $\ev_{n+1}$ is being added, then:
\begin{equation}\label{GMES_Markov_final}
	\GM(\ES, \ev) \propto \GM(\ES) \cdot \biggl(p_c(\ev) \cdot \Gamma_m(\ev) + p_u(\ev)\biggr)
\end{equation}
where at $n=0$ $\GM$ is initialized to the a-priori distribution $\GAM$.

\subsubsection{Interpretation of $\GM(\ES)$ as a set of hypotheses}
Analysis of the members formed in $\GM(\ES)$ while new evidence \evm is added reaches interesting conclusions.
\GMevm is expressed as the sum of $k$ Gaussian distributions $\mathcal{N}(\mu, \Sigma)$ over $\MS$:
\begin{equation}
	\GMev = \MSp(\m | \ev) = \sum\limits_{i=1}^{k} a_i \cdot \mathcal{N}(\mu_i, \Sigma_i) 
\end{equation}
\GMevm can be seen as set $\HipS(\ev)$ of \textbf{hypotheses}  explaining \evm, where $\mu_i$ is the hypothetical model \mm that created \evm,  $\Sigma_i$ represents the hypothesis dispersion, and $a_i$ represents its probability. Note that there should be adequate overlap between hypotheses to compensate for the fact that it is actually a continuum. 
Let $\Hip_i(\ev)$ be hypothesis $i$ in  $\HipS(\ev)$:
\begin{equation}
	\Hip_i(\ev) = a_i \cdot \mathcal{N}(\mu_i, \Sigma_i) 
\end{equation}
then
\begin{equation}
	\GMev = \sum_{\Hip \in \HipS} \Hip(\ev) 
\end{equation}
\GMevm influences $\GM(\ES)$ through  $\Gamma_m(c)$, which perform  division by $\GAM$. The quotient of two Gaussian densities is an unnormalized Gaussian density as well (see App. \ref{App_2G}). To enable that, $\GAM$ is represented by a single Gaussian distribution with large covariance matrix, resulting in $\Gamma_m(c)$ being a Gaussian mixture as well.

$\Gamma_m(\ESC)$ is the sum of all possible products of a single hypothesis for each $c \in \ESC$. As the product of Gaussian distributions is an unnormalized Gaussian distribution as well (see app. \ref{App_2G}), both $\Gamma_m(\ESC)$ and $\GM(\ES)$ are also sets of unnormalized Gaussian distribution.
We can now write  $\GM(\ES)$ as a set of hypothesis about $\ES$, $\HipS(\ES)=\{\Hip_1(\ES), \Hip_2(\ES)\cdots\} $ each resulting from a different combination of hypotheses about individual \evm, including the possibility that it is noise:
\begin{align}
\begin{split}
	\HipS(\ES) & = \{\Hip_1^1(\ev_1) , \Hip_2^1(\ev_1) \cdots \Hip_{K_1}^1(\ev_1), p_u(\ev_1) \} \\
	& 	\times \{\Hip_1^2(\ev_2) , \Hip_2^2(\ev_2)\cdots \Hip_{K_2}^1(\ev_1), p_u(\ev_2) \} \\
	& \vdots \\
	& \times  \{\Hip_1^n(\ev_n) , \Hip_2^n(\ev_n) \cdots \Hip_{K_n}^n(\ev_1), p_u(\ev_n) \}
\end{split}
\end{align}
where $\Hip_i^j(\ev_i)$ is hypothesis $i$ about $\ev_j$, $K_j$ is the number of hypotheses about $\ev_j$, and $p_u(\ev_j)$ is the probability density that this evidence is noise. Keep in mind that $\Hip(\ES)$ is an unnormalized Gaussian density. $\GM(\ES)$ can be expressed as a sum of hypothesis about the whole evidence set:
\begin{equation}\label{GM_as_Hipo}
	\GM(\ES) = \GAM \cdot \sum_{\Hip(\ES) \in \HipS(\ES)} \Hip(\ES) 
\end{equation}

\paragraph{Managing $\GM(\ES)$:}\label{GMM_WHY_AND_HOW_1}
When new evidence are added, the number of members in Eq. \ref{GM_as_Hipo} grows exponentially.
In order to keep it usable, similar components are merged, and the ones with least population ratio are removed. Component similarity is checked using  \href{https://en.wikipedia.org/wiki/Bhattacharyya_distance} {Bhattacharyya distance}\citep{Math_Bhattacharyya}. This issue is part of the implementation issues, which are still under development and will be discussed in future publications.

\subsection{Detection as {\OR} by-product}
Detection solutions are built to detect existence of an object of interest and possibly its properties, spatial and others. The detection results in either a single binary outcome, or a vector of object properties. This simple output enables rating and benchmarking by comparing the result to ground truth.

But what if the \textbf{correct} answer is not just TRUE / FALSE or a property vector? Consider the following case: upright rectangle detection encounters an occluded object where only a single corner is clearly visible. Object existence probability is not negligible, and its properties range is much more focused than its a-priory distribution. If the detection process terminates here, the result is unclear. Using {\OR} terminology, $\GM(\ES)$ is not concentrated in a single narrow range, and thus cannot yield a decisive result. However, $\GM(\ES)$ \textbf{does} contain valuable information that can be used in a larger context. Many industrial detection processes are part of a larger context and can use this information. Here are some examples:
\begin{itemize}
	\item Robotic active vision might use $\GM(\ES)$ to target the area that might complete missing information;
	\item Real-time 3D scanners can use $\GM(\ES)$ to build an approximate model of the scanned object at early stages and filter out other objects.
	\item Surveillance systems can update $\GM(\ES)$ with each incoming image to accumulate information about a tracked object.
	\item Face detection can assist person detection by sharing its $\GM(\ES)$.
\end{itemize}
\textbf{{\OR} does not equal detection:} its goal is to update $\GM(\ES)$ with new evidence and keep it as accurate as possible. Detection can be context-specific and take one of several forms:
\noindent
\begin{enumerate}%[wide, labelwidth=!, labelindent=10pt]
	\item \textbf{Find the best matches} Typically we expect one match, but in cases of low quality input or not enough evidence there might be several possible matches.
	In {\OR}'s case, these are $\GM(\ES)$ local maxima.
	\item \textbf{Calculate the portion of distribution attributed to each match}: Calculate the volume of $\GM(\ES)$ in a range $D_M$ around each match $m$ . The range is context-specific and defines the acceptable parameter range of a match. The more accurate the match has to be, the smaller $D_M$ is:
	\begin{equation}
		P_m = \int\limits_{m-D_M}^{m+D_M}\GM(\ES) \diff\MS
	\end{equation}
	$P_m$ is the probability that a match is inside range $m \pm D_M$.
	\item \textbf{Calculate the probability that match $m$ is object vs. random noise}:
	The Gaussian components of $\GM(\ES)$  comprise the hypothesis set $\HipS(\ES)$; the noise ratio of each hypothesis $\Hip(\ESC, \ESU) \in \HipS(\ES)$  is $N_R(\Hip) = \frac{\#\ESU(\Hip) }{\#\ES}$ where $\ESU(\Hip)$ is the set of noise evidence assumed by $\Hip$. The total $N_R$ will be a weighted average:
	\begin{equation}
	N_R(\m) = \frac
	{\sum\limits_{\Hip \in \HipS(\ES)} N_R(\Hip) \cdot \Hip(\m) }
	{\sum\limits_{\Hip \in \HipS(\ES)}  \Hip(\m)}
	\end{equation}
where $\Hip(\m)$ is the value of the relevant Gaussian component at \mm.
	
\end{enumerate}

\section{Test results}\label{Sec_DemoCasesResults}
Tests started at early stages during {\OR} development in order to learn its behavior and eliminate mathematical errors. 
The tests are planned to continue throughout implementation development, which is ongoing; I plan to publish a continuation article describing it. At this stage three test cases are used as described in \ref{DemoCase1} - \ref{DemoCaseLast} : upright rectangle, 4X4 grid, and spiral sector. In the future, more test cases will be added, including real-world images.

\begin{figure}[h]
	\centering
	\includegraphics[width=0.9\linewidth]{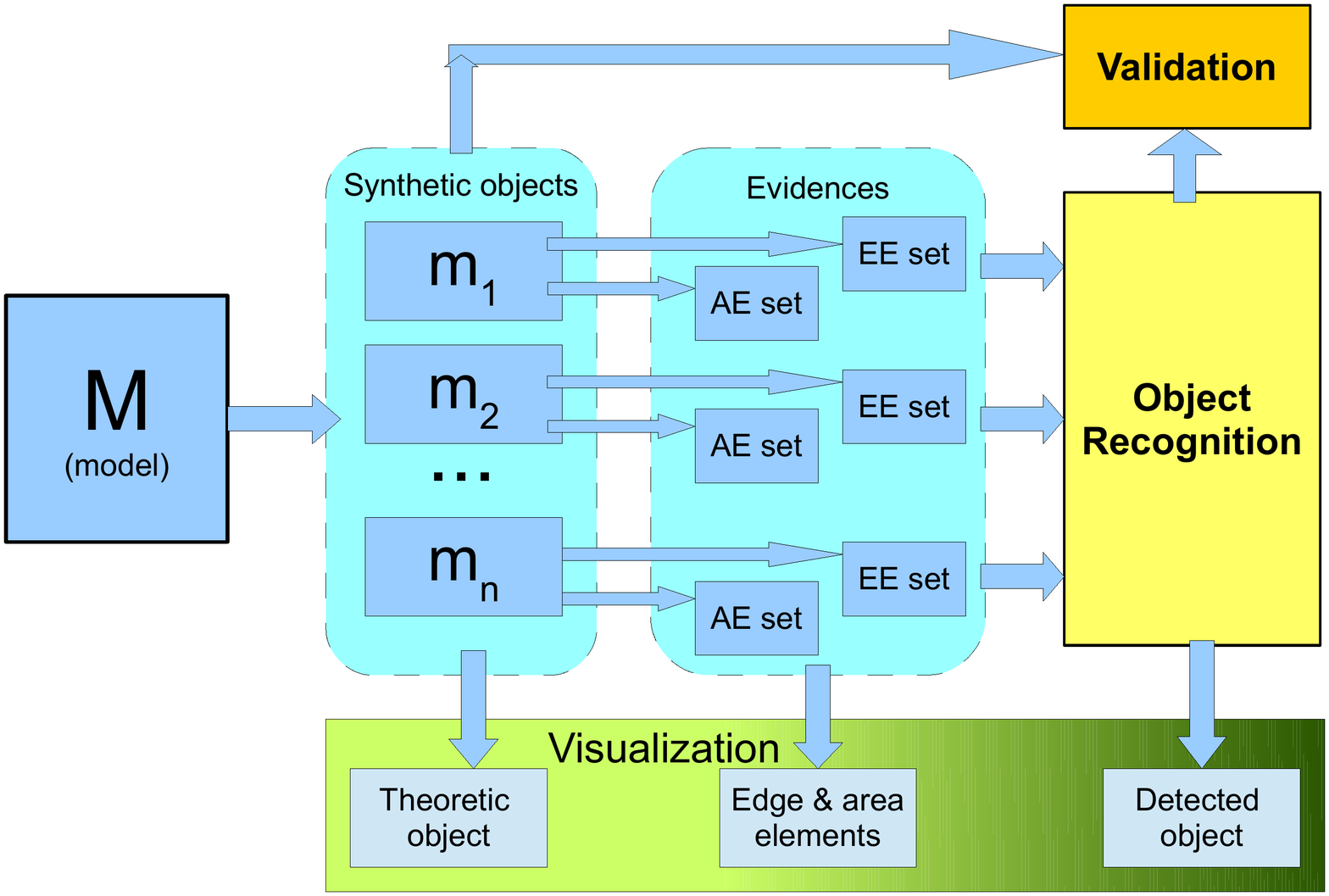}
	\caption{Overall flow of the project}
	\label{ProjectOverall1}
\end{figure}

Test policy is as follows:
for each case, a variety of objects is created programatically using a range of parameters.
The sets of evidence are created \textbf{directly} from each object model's parameters, \textbf{not from any images}. Random noise, scatter and missing coverage are introduced as well to simulate realistic scenarios. See Fig. \ref{ProjectOverall1} for a top level representation of data flow. 
This test policy, of creating the evidence programatically instead of detecting it in images, is necessary to avoid dependency on edge detection specifics or any other preprocessing steps, and focuses the research on {\OR} unique contribution. The work presented here is not about the best way to detect these specific objects, but rather about detecting objects when given a set of evidence, which may well be sub-optimal.
Processing real-world images and checking against benchmarks will be explored in future works.

The three cases were each tested by creating 25 random instances of the model, repeating detection 20 times for each instance, altogether 500 repetitions per model. In each repetition a different evidence set was created programatically with random noise and scatter according to the selected quality level. {\OR} parameters were selected to yield a convenient compromise between speed and accuracy.
%The evidence were processed in random order, updating  $\GM(\ES)$ on each new evidence. 
%This processing scheme is more challenging that most common scenarios where a set of evidence is first processed as a whole to narrow $\Phi_{\mathsf{M}}$ or filter noises. 
Samples of the results are available, including visualization of the input evidence and $\Phi_{\mathsf{M}}$ narrowing along the process at:

\noindent \href{http://www.odedcohen.com/orex/v2/index.html\#test\_results}{\textsf{http://www.odedcohen.com/orex/v2/index.html\#test\_results}}  
 
After some experimentation and parameter tuning a detection rate of $> 0.95$ was achieved for upright rectangle and 4X4 grid on low-quality input; testing of spiral sector still shows some implementation problems which have to be solved. About 70\% of the errors resulted from 30\% of the instances, indicating that the approximation error of \GMevm as GMM has several local maximum that are large enough to create slightly more error-prone domains. These are not yet quantitive results, but strong correlation was found between two implementation parameters and the detection rate:
\begin{enumerate}
	\item Maximal allowed number of Gaussian components in $\GM(\ES)$: decreasing it by 1/2 doubles the error rate. It makes sense, as this is the maximum number of hypotheses allowed at any given moment.
	\item Extending the study set parameter range beyond that of the test set decreases the error rate considerably, which indicates that \GMevm GMM approximation tends to be less effective at its margins.
\end{enumerate}

The tests so far indicate that {\OR} model represents the detection process well, which is the primary goal of this work. The implementation was kept basic and simple; it will be further improved and optimized in the next steps of this project.

\section{Discussion and the road ahead}\label{Sec_implications} 
This works presents the theory of {\OR}, and some preliminary test results on synthetic images, each containing a single object. Next development steps will deal with real-world input, and extend the model to scenes (2D or 3D) with multiple objects. The main contribution of this work is the theoretical aspect of {\OR}. 
In the long run, after sufficient testing and documentation, I plan to make it an open source project.

Since 2015 the vast majority of works in this area discuss DL and network architectures (\citet{Disc_20YearsSurvey}); they basically describe system optimization by educated architecture trial-and-error, backed-up by benchmarks.
On the contrary, {\OR} defines a system with not place for heuristic adjustments, except for implementation constraints such as merging hypotheses and limiting their number. Put simply, {\OR} calculates the match PDF over the solution space as a joint distribution of the input evidence, using basic Bayesian probability. It is simple and sound enough to render pointless any algorithmic adjustments.

{\OR} future development should adhere to this approach. For example, when detecting two instances instead of one, it is tempting to look for two salient match regions in $\MS$, as with GHT(\citep{GHT_Ballard}); this is a mistake. Two salient match ranges in $\MS$ simply imply ambiguity for a single object; instead, one should search $\MS^\textbf{2}$ for a \textbf{single} salient match range. What about the case of unknown number of objects? {\OR} next development stages will enter uncharted waters, where the only way to stay on track is by adhering to a \textbf{sound and solid probabilistic model that leaves nearly no place for heuristics}. Here are some examples of potential future fruits:
\begin{itemize}
	\item \textbf{Dynamic dimensionality:} Selection of model dimensions is one of the most crucial decisions of any object recognition system design; high dimensionality might be necessary to reliably describe the model at hand, but usually requires a more complex solution and a more extensive data gathering.
Later adjustment to model dimensionality, resulting from new information, might require extensive changes and re-training. 

{\OR} has a potential to dynamically optimize dimensionality at run-time, by analyzing $\GM(\ES)$. Each hypothesis is presented by a Gaussian component. If a group $\mathbbm{G}$ of hypotheses has a common factor $\mathcal{F}$, 
\begin{equation}
	\sum\limits^{\mathbbm{G}} \mathcal{N}_d =  \mathcal{N}_{k} \cdot \sum\limits^{\mathbbm{G}} \mathcal{N}_{(d-k)} 
\end{equation}
where $d$ is $\MS$ dimensionality, and  $\mathcal{N}_{(d-k)} , \mathcal{N}_k$ are  multivariate Gaussian distributions of dimensionality $(d-k)$ and $k$ respectively,
then $\mathcal{N}_k$ is also a common factor for all joint distributions resulting from products between $\mathbbm{G}$ members ; $\mathcal{N}_k$ expresses a set of $k$ dimensions that are irrelevant \textbf{inside this group of hypotheses}. When $\GM(\ES)$ converges to narrow match ranges, it is feasible that the hypotheses in each range share a significant common factor, and that the products of hypotheses across different match ranges are negligible. In other words, when more evidence are added the solution space can be split into several sub-spaces of lower dimensionality.

In a similar way, it is possible to extend \GMevm dimensionality, for example to cope with newly discovered latent variables while still using existing data, by selectively multiplying groups of hypotheses by common factors. 

	\item \textbf{Efficient evidence grouping:}
Primitive evidence (e.g. edge elements) are often grouped into larger ones (e.g. line segments) to create more informative evidence. However, this comes with the cost of losing some information, and the risk of sub-optimal grouping which might degrade performance in the next processing steps. 
%Correct grouping is one of the most challenging tasks of object recognition systems that use complex model, like articulated...

Using {\OR}, it is possible to reliably group evidence with minimal loss of information. A set $\mathbbm{J}$ of evidence can be grouped simply by creating its $\GM(\mathbbm{J})$ (Eq. \ref{GMES_Markov_final}); if $\mathbbm{J}$ \textbf{do} comprise a reasonable group (e.g. co-linear edge elements),  $\mathbbm{J}$ should contain several high-probability hypotheses with small covariance. For example, if the model is a triangle defined by three corners, and $\mathbbm{J}$ members are mainly EE of one of its sides, then $\GM(\mathbbm{J})$ will contain high-probability low-variance hypotheses about the two relevant corners. $\GM(\mathbbm{J})$ is used in this context as a clustering distance function.
To take full advantage of grouping, Eq. \ref{GMES_Markov_final} can be extended to use  $\GM(\mathbbm{J})$ instead of $\GMev$. There are several reasons why {\OR} can benefit from grouping:
\begin{enumerate}
	\item It can be the basis for breaking the process represented by Eq. \ref{GMES_Markov_final} into multiple parallel processes of grouping and merging.
	\item In the future when {\OR} is extended to scenes with multiple objects of various types, grouping can create groups usable by several probability spaces.
\end{enumerate}
 
\end{itemize}
These are preliminary ideas that require more R\&D, and should be treated as such.

\bibliographystyle{plainnat}
\bibliography{orex} 

\begin{thebibliography}{16}
\providecommand{\natexlab}[1]{#1}
\providecommand{\url}[1]{\texttt{#1}}
\expandafter\ifx\csname urlstyle\endcsname\relax
  \providecommand{\doi}[1]{doi: #1}\else
  \providecommand{\doi}{doi: \begingroup \urlstyle{rm}\Url}\fi

\bibitem[Mat()]{Math_Bhattacharyya}
Bhattacharyya distance.
\newblock \url{https://en.wikipedia.org/wiki/Bhattacharyya_distance}.

\bibitem[Ballard(1981)]{GHT_Ballard}
D.H. Ballard.
\newblock Generalizing the hough transform to detect arbitrary shapes.
\newblock \emph{Pattern Recognition}, 13\penalty0 (2):\penalty0 111--122, 1981.

\bibitem[Bulat and Tzimiropoulos(2016)]{ANN_RegHM_Bulat_2016}
Adrian Bulat and Georgios Tzimiropoulos.
\newblock Human pose estimation via convolutional part heatmap regression.
\newblock \emph{Lecture Notes in Computer Science}, page 717–732, 2016.
\newblock ISSN 1611-3349.
\newblock \doi{10.1007/978-3-319-46478-7_44}.
\newblock URL \url{http://dx.doi.org/10.1007/978-3-319-46478-7_44}.

\bibitem[Fischler and Bolles(1981)]{RANSAC_Fischler1981RandomSC}
M.~Fischler and R.~Bolles.
\newblock Random sample consensus: a paradigm for model fitting with
  applications to image analysis and automated cartography.
\newblock \emph{Commun. ACM}, 24:\penalty0 381--395, 1981.

\bibitem[Hubert et~al.(2005)Hubert, Rousseeuw, and Branden]{REG_Cylinder2_RPCA}
Mia Hubert, Peter Rousseeuw, and Karlien Branden.
\newblock Robpca: A new approach to robust principal component analysis.
\newblock \emph{Technometrics}, 47:\penalty0 64--79, 02 2005.
\newblock \doi{10.1198/004017004000000563}.

\bibitem[Kiryati et~al.(1991)Kiryati, Eldar, and Bruckstein]{GHT_PHT}
N.~Kiryati, Y.~Eldar, and A.M. Bruckstein.
\newblock A probabilistic hough transform.
\newblock \emph{Pattern Recognition}, 24\penalty0 (4):\penalty0 303--316, 1991.
\newblock ISSN 0031-3203.
\newblock \doi{https://doi.org/10.1016/0031-3203(91)90073-E}.
\newblock URL
  \url{https://www.sciencedirect.com/science/article/pii/003132039190073E}.

\bibitem[Li and Zhang(2013)]{CLA_Li_SURF_face_car}
Jianguo Li and Yimin Zhang.
\newblock Learning surf cascade for fast and accurate object detection.
\newblock In \emph{Proceedings of the IEEE conference on computer vision and
  pattern recognition}, pages 3468--3475, 2013.

\bibitem[Malik et~al.(2016)Malik, Arbeláez, Carreira, Fragkiadaki, Girshick,
  Gkioxari, Gupta, Hariharan, Kar, and Tulsiani]{Conc_3R}
Jitendra Malik, Pablo Arbeláez, João Carreira, Katerina Fragkiadaki, Ross
  Girshick, Georgia Gkioxari, Saurabh Gupta, Bharath Hariharan, Abhishek Kar,
  and Shubham Tulsiani.
\newblock The three r’s of computer vision: Recognition, reconstruction and
  reorganization.
\newblock \emph{Pattern Recognition Letters}, 72:\penalty0 4--14, 2016.
\newblock ISSN 0167-8655.
\newblock \doi{https://doi.org/10.1016/j.patrec.2016.01.019}.
\newblock URL
  \url{https://www.sciencedirect.com/science/article/pii/S0167865516000313}.
\newblock Special Issue on ICPR 2014 Awarded Papers.

\bibitem[Mohseni et~al.(2020)Mohseni, Zarei, and Ragan]{Conc_AeplanatoryAI}
Sina Mohseni, Niloofar Zarei, and Eric~D. Ragan.
\newblock A multidisciplinary survey and framework for design and evaluation of
  explainable ai systems, 2020.

\bibitem[Mukhopadhyay and Chaudhuri(2015)]{GHT_Survey}
Priyanka Mukhopadhyay and Bidyut~B. Chaudhuri.
\newblock A survey of hough transform.
\newblock \emph{Pattern Recognition}, 48\penalty0 (3):\penalty0 993--1010,
  2015.
\newblock ISSN 0031-3203.
\newblock \doi{https://doi.org/10.1016/j.patcog.2014.08.027}.
\newblock URL
  \url{https://www.sciencedirect.com/science/article/pii/S0031320314003446}.

\bibitem[Nurunnabi et~al.(2017)Nurunnabi, Sadahiro, and
  Lindenbergh]{REG_Cylinder2}
Abdul Nurunnabi, Yukio Sadahiro, and Roderik Lindenbergh.
\newblock Robust cylinder fitting in three-dimensional point cloud data.
\newblock \emph{ISPRS - International Archives of the Photogrammetry, Remote
  Sensing and Spatial Information Sciences}, XLII-1/W1:\penalty0 63--70, 05
  2017.
\newblock \doi{10.5194/isprs-archives-XLII-1-W1-63-2017}.

\bibitem[Toshev and Szegedy(2014)]{ANN_RegDF_Toshev_2014}
Alexander Toshev and Christian Szegedy.
\newblock Deeppose: Human pose estimation via deep neural networks.
\newblock \emph{2014 IEEE Conference on Computer Vision and Pattern
  Recognition}, Jun 2014.
\newblock \doi{10.1109/cvpr.2014.214}.
\newblock URL \url{http://dx.doi.org/10.1109/CVPR.2014.214}.

\bibitem[Tran et~al.(2015)Tran, Cao, and Laurendeau]{REG_Cylinder1}
Trung-Thien Tran, Van-Toan Cao, and Denis Laurendeau.
\newblock Extraction of cylinders and estimation of their parameters from point
  clouds.
\newblock \emph{Computers and Graphics}, 46:\penalty0 345--357, 02 2015.
\newblock \doi{10.1016/j.cag.2014.09.027}.

\bibitem[{Viola} and {Jones}(2001)]{CLA_Viola}
P.~{Viola} and M.~{Jones}.
\newblock Rapid object detection using a boosted cascade of simple features.
\newblock In \emph{Proceedings of the 2001 IEEE Computer Society Conference on
  Computer Vision and Pattern Recognition. CVPR 2001}, volume~1, pages I--I,
  2001.
\newblock \doi{10.1109/CVPR.2001.990517}.

\bibitem[Xu et~al.(1990)Xu, Oja, and Kultanen]{GHT_RHT}
Lei Xu, Erkki Oja, and Pekka Kultanen.
\newblock A new curve detection method: Randomized hough transform (rht).
\newblock \emph{Pattern Recognition Letters}, 11\penalty0 (5):\penalty0
  331--338, 1990.
\newblock ISSN 0167-8655.
\newblock \doi{https://doi.org/10.1016/0167-8655(90)90042-Z}.
\newblock URL
  \url{https://www.sciencedirect.com/science/article/pii/016786559090042Z}.

\bibitem[Zou et~al.(2019)Zou, Shi, Guo, and Ye]{Disc_20YearsSurvey}
Zhengxia Zou, Zhenwei Shi, Yuhong Guo, and Jieping Ye.
\newblock Object detection in 20 years: A survey, 2019.

\end{thebibliography}

\begin{appendices}

\section{Product and quotient of two Gaussian densities}\label{App_2G}
Let $\mathcal{N}(x|\mathbf{m}_1, \Sigma_1), \mathcal{N}(x|\mathbf{m}_2, \Sigma_2)$ be two Gaussian densities. Their product is an unnormalized Gaussian density as well:
\begin{equation}
	\begin{split}
		&\mathcal{N}(x|\mathbf{m}_1, \Sigma_1) \cdot \mathcal{N}(x|\mathbf{m}_2, \Sigma_2) \propto \mathcal{N}(x|\mathbf{m}, \Sigma)\\
		& \text{where} \\
		&\Sigma  = (\Sigma_1^{-1}+\Sigma_2^{-1})^{-1}\\
		&\mathbf{m} = \Sigma(\Sigma_1^{-1}\mathbf{m}_1 + \Sigma_2^{-1}\mathbf{m}_2),\\
		%\mathcal{Z}& = \frac{|\Sigma_2|}{|\Sigma_2-\Sigma_1|}\cdot \frac{1}{\mathcal{N}(\mathbf{m}_1|\mathbf{m}_2, \Sigma_2-\Sigma_1)}.
	\end{split}
\end{equation}
and their quotient will be
\begin{equation}
	\begin{split}
	&\frac{\mathcal{N}(x|\mathbf{m}_1, \Sigma_1)}{\mathcal{N}(x|\mathbf{m}_2, \Sigma_2)} \propto \mathcal{N}(x|\mathbf{m}, \Sigma)\\
	& \text{where} \\
 & \Sigma = (\Sigma_1^{-1}-\Sigma_2^{-1})^{-1}\\
 & \mathbf{m} = \Sigma(\Sigma_1^{-1}\mathbf{m}_1 - \Sigma_2^{-1}\mathbf{m}_2),\\
%\mathcal{Z}& = \frac{|\Sigma_2|}{|\Sigma_2-\Sigma_1|}\cdot \frac{1}{\mathcal{N}(\mathbf{m}_1|\mathbf{m}_2, \Sigma_2-\Sigma_1)}.
\end{split}
\end{equation}
Note that for the quotient, the result is a valid Gaussian distribution only if $\Sigma$ is definite positive.
\section{Symbol list}

\begin{tabular}{ |l|l| } 
	\hline
	Symbol & Meaning \\ 
	\hline
	\hline
	$\M$ &  object model (rectangle, circle...) \\ 
	\hline
	$\MS$ & model parameters space \\ 
	\hline
	$\m$ & object instance, point in a model parameter space\\ 
	\hline
	$\GM$ &  \mm distribution over $\MS$ \\ 
	\hline
	$\GMev$ & \mm distribution over $\MS$ given evidence \evm \\ 
	\hline
	$\GAM$ & \mm a-priopi distribution \\ 
	\hline
	$p(\m) , \MSp(\m)$  & \mm PDF over $\MS$ \\ 
	\hline
	\hline
	%------------------------------------------
	$\ev$ & single evidence (edge element, area patch etc.)\\ 
	\hline
	$\evS$ &  evidence space \\ 
	\hline
	$\evSe , \evSa$ &  specific evidence space \\ 
	\hline
	$p(\ev) , \evp(\ev)$ &  \evm probability density over $\evS$ \\ 
	\hline
	$p_c(\ev) , p_u(\ev)$ & \Longunderstack[c]{ $\evp(\ev)$ components: object-correlated, and noise\\  $p(\ev) = p_c(\ev) + p_u(\ev)$}  \\ 
	\hline
	\hline
	%------------------------------------------
	$\ES$ &  evidence set \\ 
	\hline
	$\ESS$ &  \Longunderstack[l]{Evidence-set space: \\ $\ESS = \evS_1 \times \evS_2... \times \evS_n = \prod_{E} \evS $} \\ 
	\hline
	$p(\ES) , \ESp(\ES)$ &  $\ES$ PDF over $\ESS$ \\ 
	\hline
	$\ESC$ &  object-correlated evidence set \\ 
	\hline
	
	$\ESCS$ & object-correlated evidence set space\\ 
	\hline
	
	$p(\ESC) , \ESCp(\ESC) $ &  $\ESC$ PDF over $\ESCS$ \\ 
	\hline
	%-- -- -  --- - - -- 
	$\ESU$ &  random-noise evidence set \\ 
	\hline
	$\ESUS$ &  random-noise evidence set space\\ 
	\hline
	
	$p(\ESU) , \ESUp(\ESU)$ & $\ESU$ PDF over $\ESUS$ \\ 
	\hline
	\hline
	%------------------------------------------
	$\obS$ &  observation space,  $\obS = \MS \times \evS$ \\ 
	\hline
	
	$\ob$ &  single observation, \{\evm , \mm\} \\ 
	\hline
	$\GOB(\ob) , \GOBo$ &  observation distribution over $\obS$ \\ 
	\hline
	%$\OBSS$ & observation-set space,  $\OBSS = \MS \times \ESS$ \\ 
	%\hline
\end{tabular}

\end{appendices}

\end{document}